\documentclass[11pt]{article}

\usepackage[margin=1.15in]{geometry}
\usepackage{amsmath,amssymb,amsthm}
\usepackage{graphicx}
\usepackage{booktabs}
\usepackage{array}
\usepackage{pifont}
\usepackage{microtype}
\usepackage[round]{natbib}
\usepackage{xcolor}
\definecolor{linkblue}{RGB}{20,60,150}
\usepackage[colorlinks=true,linkcolor=linkblue,citecolor=linkblue,urlcolor=linkblue]{hyperref}

\newtheorem{theorem}{Theorem}
\newtheorem{proposition}[theorem]{Proposition}
\newtheorem{corollary}[theorem]{Corollary}
\theoremstyle{definition}
\newtheorem{definition}{Definition}
\newtheorem{assumption}{Assumption}

\theoremstyle{remark}
\newtheorem{remark}{Remark}[section]

\newcommand{\bmix}{\beta_{\mathrm{mix}}}
\newcommand{\E}{\mathbb{E}}
\newcommand{\Prob}{\mathbb{P}}
\newcommand{\Var}{\mathrm{Var}}
\newcommand{\MSE}{\mathrm{MSE}}
\newcommand{\scheme}{\mathcal{S}}
\newcommand{\law}{\mathcal{L}}
\newcommand{\tv}[1]{\left\lVert #1\right\rVert_{\mathrm{TV}}}
\newcommand{\dist}{\operatorname{dist}}
\newcommand{\Leff}{\Lambda_{\mathrm{eff}}}
\newcommand{\filt}{\mathcal{F}}

\title{The Impossible Trinity of Time-Series Validation:\\
A Conservation Law among Training Sufficiency,\\
Test Coverage, and Temporal Causality}

\author{Jiayu Li\thanks{Email: \texttt{lijiayu2027@outlook.com}.}}

\date{September 2026}

\begin{document}

\maketitle

\begin{abstract}
Validating a model on a time series asks for three things at once: each training run should use most of the sample (\emph{sufficiency}), the test sets should together cover most of the sample (\emph{coverage}), and training data should come before test data (\emph{causality}). We prove that the three cannot be had together and price each one. Let $\alpha$ be the smallest training fraction over folds, $\beta$ the fraction of the sample covered by tests, $\Lambda$ the fraction of the sample used as training data from the future of a test point, and $\delta$ the distance from a test point to the nearest training point in its future. Every scheme on a sample of length $T$ satisfies
\[
\alpha+\beta\ \le\ 1+\Lambda
\qquad\text{and}\qquad
\alpha+\min\{\beta,\ \delta/T\}\ \le\ 1,
\]
and under $\beta$-mixing the leakage bias at a test point is at most $2M\bmix(\delta)$. In words: going beyond the causal frontier $\alpha+\beta=1$ requires training on the future; that future data must sit within $(1-\alpha)T$ of a test point; and its harm depends on its distance, not its amount. Hence expanding walk-forward is exactly the Pareto frontier of causal validation, $k$-fold cross-validation buys the most future data, and purged $k$-fold with an embargo pays in distance instead, which is cheap when the process forgets quickly but cannot repair the part of causality demanded by non-stationarity. On pure noise, shuffled 5-fold reports an information coefficient of $+0.32$, while contiguous 5-fold, using the same amount of future data, reports $+0.004$.
\end{abstract}

\section{Introduction}\label{sec:intro}

\subsection{The problem}

Given a sample of length $T$ from a time series (e.g., feature--label pairs built from stock data), we wish to evaluate a learning algorithm, or to tune its hyperparameters. Any evaluation must split the sample into one or more training/test pairs. Practitioners hold three requirements for such splits, each individually beyond reproach:

\begin{itemize}
\item \textbf{(P1) Training sufficiency.} In each single training run, the training set should be as large as possible, ideally approaching the whole sample. Rationale: the deployed model will be trained on all $T$ samples; the further the evaluation-time training size is from $T$, the further the evaluated object is from the deployed object.
\item \textbf{(P2) Test coverage.} The union of the test sets across runs should cover as much of the sample as possible. Rationale: more test points mean lower estimator variance, and more covered market regimes mean more robust conclusions.
\item \textbf{(P3) Temporal causality.} Training data should precede test data in time. When this is violated, the severity of the violation can be measured as follows: \emph{walking from the test set toward the future, the smaller the distance to the first training sample encountered, the more severe the violation.} Rationale: time series carry serial dependence and look-ahead information; training on data in the immediate future of a test point effectively feeds the test point's own information to the model.
\end{itemize}

The four standard schemes each trade off these demands differently (Table~\ref{tab:folklore}). None achieves all three. The starting point of this paper is the claim that \emph{this is not a failure of ingenuity in scheme design, but a mathematical law} --- and that the law has a clean form: one inequality among three quantities measured on a common scale, overlaid with a statistical layer whose exchange rate is set by the memory of the process.

\begin{table}[ht]
\centering
\caption{The folklore trade-off among the four standard validation schemes.}
\label{tab:folklore}
\small
\begin{tabular}{lccc}
\toprule
Scheme & P1 sufficiency & P2 coverage & P3 causality \\
\midrule
Walk-forward (rolling/expanding) & \ding{55} (early folds starve) & partial & \checkmark \\
$k$-fold cross-validation & \checkmark & \checkmark & \ding{55} \\
Last-block hold-out & \checkmark & \ding{55} & \checkmark \\
Purged $k$-fold + embargo & partial & partial & partial (balanced) \\
\bottomrule
\end{tabular}
\end{table}

\subsection{Contributions}

\begin{enumerate}
\item \textbf{Formalization} (Section~\ref{sec:framework}). The three demands become computable coordinates of a scheme: the worst-fold training fraction $\alpha$, the coverage $\beta$, and, for causality, the anti-causal mass $\Lambda$ (how much future data is used) together with the anti-causal margin $\delta$ (how close it sits to the tests --- exactly the distance of P3).
\item \textbf{The main theorem and its combinatorics} (Section~\ref{sec:main}). Every scheme satisfies the \emph{ledger} $\alpha+\beta\le1+\Lambda$, the \emph{proximity constraint} $\delta\le(1-\alpha)T$ whenever $\Lambda>0$, hence the \emph{trinity inequality} $\alpha+\min\{\beta,\delta/T\}\le1$; and under $\beta$-mixing the leakage bias at a test point is at most $2M\bmix(\delta)$, which proves that the distance of P3 is the right severity statistic and identifies the mixing coefficient as its exchange rate (Theorem~\ref{thm:main}; Theorem~\ref{thm:ipm} restates the exchange rate in an arbitrary integral probability metric). Consequently strict causality forces $\alpha+\beta\le1$, expanding walk-forward is exactly this frontier, $k$-fold attains the ledger with equality, and even the \emph{average} training fraction of a fully covering causal scheme is at most about one half (Theorem~\ref{thm:mean}).
\item \textbf{The price of each vertex} (Section~\ref{sec:statistical}). The sufficiency gap costs a pessimistic learning-curve bias (Proposition~\ref{prop:learning}), the coverage gap a variance of order $\sigma^2/(\beta T)$ plus an irreparable worst-case regime error (Proposition~\ref{prop:coverage}), and the three prices cannot vanish together (Corollary~\ref{cor:impossible}). The hardness of the trinity is the memory of the process (Section~\ref{sec:memory}): it vanishes in the i.i.d.\ limit, an embargo buys causality back under finite memory at a sample cost of $O(m(2H+h)/T)$, and the non-stationary part is not redeemable.
\end{enumerate}
Section~\ref{sec:schemes} places the standard schemes in these coordinates, discusses tuning, and gives a minimal numerical illustration of the volume/severity distinction; Section~\ref{sec:practice} condenses the results into practical guidance. All non-trivial proofs are collected in Appendix~\ref{app:proofs}.

\section{Formal framework}\label{sec:framework}

\subsection{Data and validation schemes}

Index the sample by time, $[T]=\{1,2,\dots,T\}$. Each index $t$ carries a sample $W_t=(X_t,Y_t)$: the feature $X_t$ is observable at time $t$, and the label $Y_t$ \emph{resolves} at time $t+H$, where $H\ge0$ is the \emph{label horizon} (e.g., $Y_t$ is the forward return over the next $H$ periods).

\begin{definition}[Validation scheme]\label{def:scheme}
A validation scheme is a finite collection
\[
\scheme=\{(R_i,E_i)\}_{i=1}^{m},\qquad R_i,E_i\subseteq[T],\quad R_i\cap E_i=\varnothing,\quad E_i\neq\varnothing,
\]
where fold $i$ trains on $R_i$ and evaluates on $E_i$. The scheme outputs the weighted average of fold test losses, $\hat L(\scheme)=\sum_i w_i\hat L_i$ with $w_i\ge0$, $\sum_i w_i=1$ (typically weighted by $|E_i|$).
\end{definition}

\subsection{The three quantities}

\begin{definition}[Training sufficiency]\label{def:alpha}
The worst-fold and average training fractions are
\[
\alpha(\scheme)=\min_i\frac{|R_i|}{T},\qquad
\bar\alpha(\scheme)=\sum_i w_i\frac{|R_i|}{T},
\]
the average being taken with the weights $w_i$ that define $\hat L(\scheme)$, so that $\bar\alpha T$ is the mean training size behind the reported estimate; with equal-length test blocks weighted by $|E_i|$ it is the plain average $\frac1m\sum_i|R_i|/T$.
\end{definition}
P1 demands $\alpha\to1$. We take the worst fold because an evaluation is only as strong as its weakest link; the average version is treated separately in Theorem~\ref{thm:mean}.

\begin{definition}[Test coverage]\label{def:beta}
$\displaystyle \beta(\scheme)=\frac{\bigl|\bigcup_i E_i\bigr|}{T}$.
\end{definition}
P2 demands $\beta\to1$.

\begin{definition}[Causality, anti-causal mass, and margins]\label{def:causal}
Write $\dist(t,S)=\min_{s\in S}|s-t|$ for the distance from an index $t$ to a set $S\subseteq[T]$, with $\dist(t,\varnothing)=+\infty$. Fold $i$ is \emph{strictly causal} if $\max R_i<\min E_i$ (all training precedes all testing); the scheme is strictly causal if every fold is. For a test point $t\in E_i$ of a general fold, split the training set into the \emph{past} and \emph{future training sets} of $t$,
\[
P_i(t)=R_i\cap[1,t),\qquad F_i(t)=R_i\cap(t,T],
\]
so that $R_i=P_i(t)\cup F_i(t)$ (as $t\notin R_i$). Causality is violated at $t$ exactly when $F_i(t)\neq\varnothing$, and the violation has two measures, the size of $F_i(t)$ and its distance from $t$:
\begin{itemize}
\item the \emph{anti-causal mass} (volume of the violation):
\[
\Lambda_i=\max_{t\in E_i}\frac{|F_i(t)|}{T}=\frac{\bigl|R_i\cap(\min E_i,T]\bigr|}{T},\qquad
\Lambda(\scheme)=\max_i\Lambda_i,
\]
the maximum being attained at the earliest test point because $F_i(t)$ shrinks as $t$ grows;
\item the \emph{pointwise anti-causal margin} (distance of the violation):
\[
\delta_i(t)=\dist\bigl(t,F_i(t)\bigr)=\min\{\,r-t:\ r\in R_i,\ r>t\,\}\in\{1,2,\dots\}\cup\{+\infty\},
\]
i.e.\ \emph{the distance to the first training point encountered walking from $t$ into the future}; the fold margin is $\delta_i=\min_{t\in E_i}\delta_i(t)$ and the scheme margin $\delta(\scheme)=\min_i\delta_i$. Symmetrically, the \emph{past margin} is $\delta_i^-(t)=\dist\bigl(t,P_i(t)\bigr)$, and $\dist(t,R_i)=\min\{\delta_i^-(t),\delta_i(t)\}$ is the \emph{two-sided margin}.
\end{itemize}
\end{definition}
Strict causality is equivalent to $\Lambda=0$ and to $\delta=+\infty$: both say that $F_i(t)=\varnothing$ at every test point of every fold. When $E_i$ is a contiguous interval with all training outside it, $\delta_i$ is attained at the right end of the interval --- the phrasing of P3 (``walking from the test set toward the future''); Definition~\ref{def:causal} is its pointwise generalization. The two-sided margin $\dist(t,R_i)$ is the exclusion radius of hv-block cross-validation \citep{racine2000}, which Section~\ref{sec:illustration} sweeps.

\begin{remark}[Why two violation measures]\label{rem:twomeasures}
$\Lambda$ and $\delta$ are two functionals of one object, the future training set $F_i(t)$ of a test point: its cardinality and its distance from $t$. $\Lambda$ measures \emph{how much} future data is used; $\delta$ measures \emph{how close} it sits. Theorem~\ref{thm:main} shows that the combinatorial ledger counts the cardinality, that a non-empty $F_i(t)$ forces a bound on the distance, and that statistical harm counts only the distance (converted through the dependence structure of the process).
\end{remark}

\begin{remark}[Causality as adaptedness]\label{rem:adapted}
Let $\filt_t=\sigma(W_s:s\le t)$ be the natural filtration of the sample (stated for $H=0$; for $H>0$ every index shifts by $H$, as in Remark~\ref{rem:horizon}). Fold $i$ is strictly causal exactly when $R_i\subseteq[1,\min E_i-1]$, in which case its trained model $\mathcal A(W_{R_i})$ is $\filt_{\min E_i-1}$-measurable, hence $\filt_{t-1}$-measurable at every test point $t\in E_i$: the fold is an \emph{adapted} predictor, and its test losses are those of a prequential evaluation in the sense of \citet{dawid1984}. Two consequences need no mixing assumption. First, $\ell_t-\E[\ell_t\mid\filt_{t-1}]$ is a martingale-difference sequence along each test block, so the causal frontier of Corollary~\ref{cor:frontier} is exactly the class of schemes to which martingale concentration and anytime-valid confidence sequences \citep{ramdas2023} apply directly. Second, an anti-causal fold is a non-adapted predictor, and Theorem~\ref{thm:main}(d) measures its non-adaptedness by the dependence between $\filt_t$ and the future $\sigma$-algebra $\sigma(W_s:s\ge t+\delta_i(t))$: the margin is the lag at which that dependence is evaluated.
\end{remark}

\subsection{Label horizon and informational order}\label{sec:horizon}

When $H>0$, index order is not information order: the label of sample $s$ resolves only at $s+H$, so at prediction time $t$ the \emph{usable} training samples are $\{s: s+H\le t\}$. Even walk-forward must therefore keep a gap of $H$ between training set and test block, or it trains on labels not yet resolved at prediction time --- a look-ahead bias of the same origin as ``using future data''. Symmetrically, training samples within $H$ \emph{before} a test block have label windows $[s,s+H]$ overlapping the test label windows, again constituting leakage; deleting them is the original motivation of purging \citep{lopezdeprado2018}.

Formally we absorb all of this into the dependence structure of the augmented sample process $W_t=(X_t,Y_t)$: even if raw returns are i.i.d., $H$-period overlapping labels make $W$ strongly dependent at lags $d<H$ and independent at $d\ge H$ (Remark~\ref{rem:overlap}). The combinatorial statements are made for $H=0$; Remark~\ref{rem:horizon} gives the $O(mH/T)$ correction for $H>0$.

\subsection{The dependence structure of the process}\label{sec:mixing}

The statistical part of the main theorem needs one assumption on the data-generating process: a uniform bound on the dependence between past and future. Stationarity is \emph{not} needed for it; it enters only later, when fold estimates are identified with a learning curve (Assumption~\ref{ass:learning}).

\begin{assumption}\label{ass:mixing}
The augmented sample process $(W_t)_{t\in\mathbb Z}$ is absolutely regular ($\beta$-mixing) with coefficients $\bmix(d)\downarrow0$, uniformly in time: for every $a$ and $d\ge1$,
\[
\bigl\lVert\,\law\bigl(W_{\le a},W_{\ge a+d}\bigr)-\law(W_{\le a})\otimes\law(W_{\ge a+d})\bigr\rVert_{\mathrm{TV}}\ \le\ \bmix(d),
\]
where $W_{\le a}=(W_s)_{s\le a}$, $W_{\ge a+d}=(W_s)_{s\ge a+d}$, and $\tv{\mu-\nu}=\sup_A|\mu(A)-\nu(A)|$. We set $\bmix(+\infty)=0$, so that $\bmix(\delta_i(t))$ is defined for causal folds.
\end{assumption}

Assumption~\ref{ass:mixing} is the total-variation case of a more general notion. For a class $\Psi$ of measurable functions of two path segments, the \emph{integral probability metric} generated by $\Psi$ \citep{muller1997} is $d_\Psi(P,Q)=\sup_{\psi\in\Psi}|\E_P\psi-\E_Q\psi|$, and the \emph{$\Psi$-dependence coefficient} of the process is
\begin{equation}\label{eq:psidep}
\beta_\Psi(d)\ =\ \sup_a\ d_\Psi\Bigl(\law\bigl(W_{\le a},W_{\ge a+d}\bigr),\ \law(W_{\le a})\otimes\law(W_{\ge a+d})\Bigr),\qquad \beta_\Psi(+\infty)=0 .
\end{equation}
Functions bounded by $M$ give $\beta_\Psi=2M\bmix$; $1$-Lipschitz functions for a metric on the path space give a Wasserstein-type coefficient of the kind introduced by \citet{dedecker2005} and \citet{wu2005}, which decays for processes that are not mixing at all, such as the autoregressions with discrete innovations of \citet{andrews1984}; and the loss maps of a fixed algorithm give a \emph{discrepancy} in the sense of \citet{kuznetsov2015,kuznetsov2020}. Theorem~\ref{thm:ipm} states the exchange rate at this level of generality; the main theorem uses the total-variation case, which asks nothing of the algorithm.

Throughout, losses are bounded, $|\ell|\le M$. For a fold $i$ and a test point $t\in E_i$, write $R^-=P_i(t)$ and $R^+=F_i(t)$ for the past and future training sets of Definition~\ref{def:causal}, so that $R^-\subseteq[1,t-\delta_i^-(t)]$ and $R^+\subseteq[t+\delta_i(t),T]$, and write
\[
\ell_t=\ell\bigl(\mathcal A(W_{R^-},W_{R^+}),\,W_t\bigr)
\]
for the fold model's loss at $t$. Two \emph{leak-free references} will be compared with $\ell_t$: $\tilde\ell_t$ is the same loss with $W_{R^+}$ replaced by a copy equal in distribution but independent of $\sigma(W_s:s\le t)$ (the anti-causal training data is decoupled from everything the test point can see); $\ell^\circ_t$ is the same loss with $(W_{R^-},W_t,W_{R^+})$ replaced by three mutually independent copies with the same marginal laws (all leakage on both sides is removed).

\section{The main theorem and its combinatorial consequences}\label{sec:main}

\begin{theorem}[The impossible trinity]\label{thm:main}
Let $\scheme$ be any validation scheme on $[T]$, with $\alpha,\beta,\Lambda,\delta$ as in Definitions~\ref{def:alpha}--\ref{def:causal}.
\begin{enumerate}
\item[(a)] (\emph{Ledger.}) $\alpha+\beta\le1+\Lambda$. Equality holds if and only if the test union is a suffix, $\bigcup_iE_i=[p,T]$, and some fold $i$ testing at $p=\min E_i$ trains on the entire prefix ($R_i\supseteq[1,p-1]$), is a worst fold ($|R_i|=\alpha T$), and carries the maximal anti-causal mass ($\Lambda_i=\Lambda$).
\item[(b)] (\emph{Proximity.}) Every fold that uses any future data ($\Lambda_i>0$) has $\delta_i\le T-|R_i|$. Hence if $\Lambda>0$ --- in particular whenever $\alpha+\beta>1$, by (a) --- then $\delta\le(1-\alpha)T$: some fold trains on a point at most $(1-\alpha)T$ steps after one of its own test points.
\item[(c)] (\emph{Trinity.}) Consequently, with the convention $\delta/T=+\infty$ for strictly causal schemes,
\[
\alpha+\min\{\beta,\ \delta/T\}\ \le\ 1 .
\]
Equality with $\delta=+\infty$ is attained at every point $(\alpha,\beta)=(1-\beta,\beta)$ of the causal frontier by the expanding walk-forward schemes of Corollary~\ref{cor:frontier}.
\item[(d)] (\emph{Exchange rate.}) Under Assumption~\ref{ass:mixing}, for every fold $i$ and test point $t\in E_i$, with $\delta^+=\delta_i(t)$ and $\delta^-=\delta_i^-(t)$,
\[
\bigl|\,\E[\ell_t]-\E[\tilde\ell_t]\,\bigr|\ \le\ 2M\,\bmix(\delta^+),
\qquad
\bigl|\,\E[\ell_t]-\E[\ell^\circ_t]\,\bigr|\ \le\ 2M\bigl[\bmix(\delta^-)+\bmix(\delta^+)\bigr].
\]
Averaging over the fold, $\bigl|\E[\hat L_i]-\E[\tilde L_i]\bigr|\le\frac{2M}{|E_i|}\sum_{t\in E_i}\bmix\bigl(\delta_i(t)\bigr)$, and similarly for the two-sided reference.
\end{enumerate}
\end{theorem}

Parts (a)--(c) carry \emph{no statistical assumptions}: they are facts about splitting a finite totally ordered set, and hold for any data, any model, and any loss. Part (d) is the only place where the process enters, and it enters through a single number, $\bmix(\delta)$.

The theorem should be read as a price list. Part (a) says that every unit of $\alpha+\beta$ above the causal frontier $\alpha+\beta=1$ must be paid for with at least one unit of anti-causal training mass. Part (b) says that the mass so bought cannot be parked far away: it must sit within $(1-\alpha)T$ of a test point, so that \emph{distance is paid out of sufficiency, one sample per unit of margin}. Part (c) puts the three coordinates on one scale: sufficiency, coverage and (normalized) margin cannot all be close to one --- the impossible trinity in one line. Part (d) says that the harm of the mass is not its volume but its distance, converted at the mixing coefficient of the process. The combination of (b) and (d) is the whole story of purged $k$-fold: it buys distance instead of volume, and distance is cheap exactly when the process forgets quickly.

The intuition behind (a): on the totally ordered time axis, \emph{the earlier a test point sits, the longer the suffix it sterilizes as unusable-for-training in its fold}. Coverage demands the existence of early test points; sufficiency forbids the existence of small training sets; the two compete for the same resource --- prefix measure --- and the only way out is to train on the suffix, which is anti-causal. The resemblance to the Mundell--Fleming trilemma \citep{mundell1963} and the CAP theorem \citep{gilbert2002} is more than rhetorical: all three are conservation phenomena in which a total order or consistency constraint locks two resources into a single budget.

\begin{figure}[tb]
\centering
\includegraphics[width=0.72\textwidth]{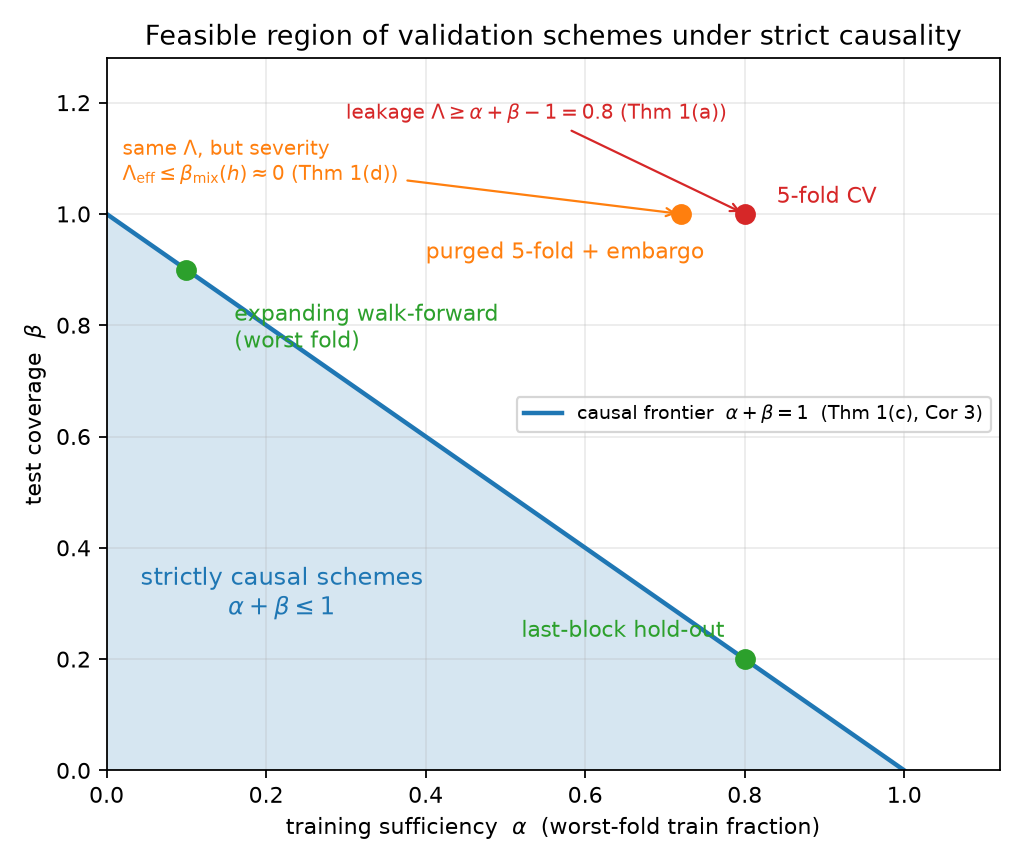}
\caption{The feasible region of strictly causal schemes, $\alpha+\beta\le1$ (Theorem~\ref{thm:main}(c)), and the coordinates of the standard schemes. $k$-fold lies outside the region and must pay $\Lambda\ge\alpha+\beta-1$ (Theorem~\ref{thm:main}(a)); purged $k$-fold has the same $\Lambda$ but its effective leakage is pushed to $\bmix(h)\approx0$ by the embargo (Theorem~\ref{thm:main}(d)), the embargo being deducted from $\alpha$ (Theorem~\ref{thm:main}(b)). The coordinates of the non-causal schemes are schematic ($k=5$; the embargo is exaggerated for visibility).}
\label{fig:frontier}
\end{figure}

\subsection{Combinatorial consequences}\label{sec:combinatorial}

\begin{corollary}[Conservation law]\label{cor:conservation}
If $\scheme$ is strictly causal, then $\alpha(\scheme)+\beta(\scheme)\le1$.
\end{corollary}

This is Theorem~\ref{thm:main}(a) with $\Lambda=0$, or (c) with $\delta=+\infty$. The equality case of (a) describes the frontier: a strictly causal scheme with $\alpha+\beta=1$ has test union $(\alpha T,T]$ and a fold that trains on exactly the prefix $[1,\alpha T]$ and tests at $\alpha T+1$.

\begin{corollary}[Attainability and Pareto frontier]\label{cor:frontier}
Let $\alpha,\beta\ge0$ with $\alpha+\beta\le1$ and $\alpha T,\beta T$ integers, $\beta T\ge1$. Then there is a strictly causal scheme attaining $(\alpha',\beta)$ with $\alpha'\ge\alpha$; moreover every point of the frontier $\alpha+\beta=1$ is attained by the following \emph{expanding walk-forward} scheme with any number $1\le m\le\beta T$ of blocks: partition $((1-\beta)T,\,T]$ into $m$ consecutive test blocks $E_1<\dots<E_m$ and set $R_i=[1,\min E_i-1]$. Conversely, every strictly causal scheme is weakly dominated in $(\alpha,\beta)$ by an expanding walk-forward scheme.
\end{corollary}

The proof is immediate: the construction has worst fold $R_1=[1,(1-\beta)T]$, so it sits at $(1-\beta,\beta)$, and by Corollary~\ref{cor:conservation} no strictly causal scheme with coverage $\beta$ has $\alpha>1-\beta$.

Corollary~\ref{cor:frontier} upgrades walk-forward from ``one scheme among many'' to \emph{the entire Pareto frontier of strictly causal validation}: the initial-window fraction $w$ slides along the frontier, tracing $(\alpha,\beta)=(w,\,1-w)$. Last-block hold-out is the member $m=1$ of the same family: in the coordinates $(\alpha,\beta)$ alone, a hold-out with test fraction $\beta$ and a walk-forward with coverage $\beta$ are indistinguishable --- both sit at $(1-\beta,\beta)$, because the worst fold of the walk-forward is exactly the hold-out. The folklore of Table~\ref{tab:folklore} (``hold-out: sufficient but poor coverage'', ``walk-forward: covers but starves'') therefore only records that hold-out is customarily run with small $\beta$ and walk-forward with large $\beta$. What genuinely separates the two at equal coverage is the \emph{average} training fraction, the subject of the next theorem: the hold-out has $\bar\alpha=1-\beta$, whereas an $m$-block walk-forward with equal blocks has $\bar\alpha=1-\frac{m+1}{2m}\beta\to1-\beta/2$.

\begin{theorem}[Mean-sufficiency version]\label{thm:mean}
Let $\scheme$ be strictly causal with pairwise disjoint test sets that are intervals of equal length $b=\beta T/m$, weighted by $|E_i|$ (so $w_i=1/m$). Then
\[
\bar\alpha(\scheme)+\frac{m+1}{2m}\,\beta(\scheme)\ \le\ 1,
\]
with equality attained by the expanding walk-forward scheme of Corollary~\ref{cor:frontier} with equal blocks. In particular, when the test blocks tile the whole sample ($\beta=1$), $\bar\alpha\le\frac{m-1}{2m}<\frac12$: \emph{even measured by average training size, a fully covering causal scheme uses at most half the data}.
\end{theorem}

At $\beta=1$ the first block has no training data at all ($\alpha=0$): full coverage is the degenerate endpoint of causal validation, and the theorem says that averaging over folds does not rescue it.

\begin{remark}[$k$-fold is the extremal buyer]\label{rem:kfoldoptimal}
Contiguous-block $k$-fold cross-validation has $\alpha=\frac{k-1}{k}$, $\beta=1$, $\Lambda=\frac{k-1}{k}$ and attains Theorem~\ref{thm:main}(a) with equality (the fold testing on the first block trains on everything after it): it spends exactly the minimum $\Lambda$ that its $(\alpha,\beta)$ requires, not a unit more --- this is $k$-fold's combinatorial optimality. All of its problems live in part (d): the $\Lambda$ it buys consists entirely of violations at margin $\delta=1$, and by part (b) it could not have pushed that margin beyond $T/k$ without giving up sufficiency.
\end{remark}

\begin{remark}[Sanitization is deducted from sufficiency]\label{rem:horizon}
Part (b) is the combinatorial half of the purged-$k$-fold story: any scheme above the causal frontier that wants its anti-causal mass at margin $\ge h$ must accept $\alpha\le1-h/T$, and with $k$ blocks the affordable margin is at most about $T/k$. The label horizon acts in the same way. For $H>0$, usable training requires $s+H\le\min E_i$, so Corollary~\ref{cor:conservation} tightens to $\alpha+\beta\le1-H/T$, and in Theorem~\ref{thm:main}(a) and Theorem~\ref{thm:mean} each train/test boundary loses a sanitization zone of length $O(H)$, a total correction of $O(mH/T)$. Both effects are the quantitative form of the folklore ``the finer you slice, the less room you have for an embargo and the more sanitization overhead you pay'' (Section~\ref{sec:memory}(b)).
\end{remark}

\section{Statistical consequences: the price of each vertex}\label{sec:statistical}

The combinatorial parts answer ``which combinations are feasible''; the statistical layer answers ``what is paid at each unsatisfied vertex''. The target of estimation is the \emph{deployment risk} $L(T)$: the expected loss of the model trained on all $T$ samples on a fresh test point, i.e.\ the learning curve of Assumption~\ref{ass:learning} evaluated at $n=T$. The scheme outputs $\hat L(\scheme)$, and we care about its bias and variance as an estimator of $L(T)$.

\subsection{The price of causality: the severity law}\label{sec:leakage}

Theorem~\ref{thm:main}(d) is the price of the causality vertex. Its first inequality isolates the anti-causal part of the training set and is the statement about P3; the second removes the leakage on both sides and produces a genuinely leak-free reference. The part has three layers of meaning:
\begin{enumerate}
\item \textbf{The severity measure of P3 is proven correct.} The bound depends on the anti-causal structure \emph{only} through $\delta^+$, the distance of P3: the coupling grabs all future training data in one stroke, so the bound sees only the nearest point, and the exchange rate of that distance is $\bmix(\cdot)$.
\item \textbf{Volume is harmless; proximity is harmful.} The same $\Lambda$ placed beyond $\delta^+\ge h$ does harm at most $2M\bmix(h)$; placed adjacent to the tests it is priced at $\bmix(1)$, typically orders of magnitude larger. Accordingly define the \emph{effective leakage}
\[
\Leff(\scheme)\ =\ \max_i\ \frac{1}{|E_i|}\sum_{t\in E_i}\bmix\bigl(\delta_i(t)\bigr)\ \le\ \bmix\bigl(\delta(\scheme)\bigr),
\]
the fold-averaged dependence coefficient at the margins, which is the quantity statistics actually needs controlled; $\Lambda$ is merely its support volume.
\item \textbf{The decoupled reference connects to the sufficiency analysis.} $\ell^\circ_t$ is the loss of a model trained on $|R_i|$ samples and evaluated at an independent test point; its expectation is the leak-free risk of the fold, which Assumption~\ref{ass:learning} below identifies with the learning curve $L(|R_i|)$. Leakage bias and learning-curve bias therefore add in the same coordinate system, and the two terms of the two-sided bound are exactly the two sanitization moves of Remark~\ref{rem:purgeembargo}.
\end{enumerate}

The total-variation bound is the worst case over all algorithms and losses. Denominating the exchange rate in the metric that matches the algorithm gives a sharper price at no extra cost in the proof.

\begin{theorem}[Exchange rate in an arbitrary metric]\label{thm:ipm}
Fix a fold $i$ and a test point $t\in E_i$ with $\delta^+=\delta_i(t)$, and let $\psi_t$ be the map $(W_{\le t},W_{\ge t+\delta^+})\mapsto\ell_t$, which exists because $(W_{P_i(t)},W_t)$ is a function of $W_{\le t}$ and $W_{F_i(t)}$ a function of $W_{\ge t+\delta^+}$. If $\psi_t\in\Psi$, then
\[
\bigl|\E[\ell_t]-\E[\tilde\ell_t]\bigr|\ \le\ \beta_\Psi(\delta^+).
\]
In particular: (i) $\Psi=\{\psi:|\psi|\le M\}$ recovers the first inequality of Theorem~\ref{thm:main}(d); (ii) if the loss composed with the algorithm is $M$-Lipschitz in the data for a metric on the path space, the bound holds with $M$ times the corresponding Wasserstein coefficient, whether or not the process is mixing; (iii) if $\Psi$ is the class of loss maps of the algorithm itself, $\beta_\Psi$ is the smallest exchange rate valid for that algorithm on every fold, and it is small for algorithm--process pairs that cannot exploit the leak. The validity of $k$-fold for autoregressions with uncorrelated errors \citep{bergmeir2018} is therefore not an exception to Theorem~\ref{thm:main}(d) but an instance of a small $\beta_\Psi$.
\end{theorem}

The proof is one line. $\E[\ell_t]=\E_P\psi_t$ with $P=\law(W_{\le t},W_{\ge t+\delta^+})$, and $\E[\tilde\ell_t]=\E_Q\psi_t$ with $Q=\law(W_{\le t})\otimes\law(W_{\ge t+\delta^+})$, because the leak-free reference replaces $W_{R^+}$ by a copy that is independent of $\filt_t$ and equal in law; hence $|\E[\ell_t]-\E[\tilde\ell_t]|\le d_\Psi(P,Q)\le\beta_\Psi(\delta^+)$ by \eqref{eq:psidep} with $a=t$, $d=\delta^+$. The two-sided inequality of Theorem~\ref{thm:main}(d) generalizes in the same way whenever $\Psi$ is closed under fixing one argument, since $d_\Psi$ is a pseudo-metric and the triangle-inequality step of Appendix~\ref{app:severity} goes through verbatim. Theorem~\ref{thm:main}(d) keeps the total-variation form because it is the only case that requires nothing of the algorithm, and because Berbee's coupling gives it a pathwise meaning (the future training data can be replaced by an independent copy on an event of probability $1-\bmix(\delta^+)$) that the expectation bound alone does not carry.

\begin{remark}[The exchange rate in bits]\label{rem:bits}
Let $I(d)=\sup_a I\bigl(W_{\le a};W_{\ge a+d}\bigr)$ be the mutual information (in nats) between the past and the $d$-separated future. Pinsker's inequality gives $\bmix(d)\le\sqrt{I(d)/2}$, so Theorem~\ref{thm:main}(d) can be restated as $|\E[\ell_t]-\E[\tilde\ell_t]|\le M\sqrt{2I(\delta^+)}$: the harm of a violation is bounded by the square root of the information that the future training set can carry about the test point. This is the form in which the same coupling appears in the information-theoretic generalization bounds of \citet{russo2016} and \citet{xu2017}, with the margin playing the role of an information budget.
\end{remark}

\begin{remark}[Purging and embargo unified]\label{rem:purgeembargo}
An embargo lifts $\delta_i(t)$ from $1$ to $h$; purging deletes training samples within $H$ of the test block on account of the label horizon, lifting $\delta_i^-(t)$ (and, on the future side, adding $H$ to the embargo). They are the same move --- ``pay distance until the decay of the mixing coefficient takes over'' --- applied to the two sides of the test block, and the two-sided bound of Theorem~\ref{thm:main}(d) prices them additively.
\end{remark}

\begin{remark}[Overlapping labels give memory to i.i.d.\ returns]\label{rem:overlap}
If raw returns are i.i.d.\ and $Y_t=\sum_{j=1}^{H}r_{t+j}$, then $\mathrm{corr}(Y_t,Y_{t+d})=\frac{(H-d)_+}{H}$: the dependence of the augmented process decays linearly within $d<H$ and cuts off to zero at $d\ge H$. The illustration of Section~\ref{sec:illustration} measures exactly this curve.
\end{remark}

\begin{remark}[Symmetry and asymmetry of direction]\label{rem:symmetry}
The two-sided bound of Theorem~\ref{thm:main}(d) treats the past and the future margins identically, and under strict stationarity the leak-free reference $\ell^\circ_t$ means the same thing on both sides: pure statistical leakage is \emph{time-symmetric} --- the distant past and the distant future are equally harmless. The genuine asymmetry of the causal direction has two sources: (i) information timing --- labels resolve in the future (Section~\ref{sec:horizon}); (ii) non-stationarity --- see Section~\ref{sec:memory}: under distribution drift, an evaluation that trains after testing answers ``could this model family have fitted that era'', not ``could you have made that money at the time''.
\end{remark}

\subsection{The price of sufficiency: learning-curve bias}\label{sec:sufficiency}

\begin{assumption}\label{ass:learning}
The process is strictly stationary. Let $L(n)$ denote the expected loss on an independent test point of the algorithm trained on $n$ samples from the process (the learning curve), assumed non-increasing in $n$ and insensitive to the internal arrangement of the training sample. The scheme is strictly causal with train/test gaps exceeding the dependence scale, so that fold estimates satisfy $\E[\hat L_i]=L(|R_i|)$ (by Theorem~\ref{thm:main}(d), the error in this identification is at most $2M\bmix(\text{gap})$, which we neglect).
\end{assumption}

\begin{proposition}[Learning-curve bias]\label{prop:learning}
Under Assumption~\ref{ass:learning},
\[
L(n_{\max})-L(T)\ \le\ \E[\hat L(\scheme)]-L(T)\ \le\ L(\alpha T)-L(T),
\]
where $n_{\max}=\max_i|R_i|$. The bias is non-negative (\emph{systematic pessimism}), with its upper bound controlled by the sufficiency gap $\alpha$; if $L$ is convex, additionally $\E[\hat L]-L(T)\ge L(\bar n)-L(T)$ with $\bar n=\sum_iw_i|R_i|=\bar\alpha T$, and by Theorem~\ref{thm:mean} a fully covering causal scheme has $\bar n\lesssim T/2$.
\end{proposition}

The proposition is immediate: $\E[\hat L]=\sum_iw_iL(|R_i|)$ with $\alpha T\le|R_i|\le n_{\max}\le T$, monotonicity gives both bounds, and Jensen's inequality gives the convex refinement. Pessimistic bias sounds ``safe'', but it is not neutral for \emph{model comparison and tuning}: learning curves of different hyperparameters have different slopes (complex models' $L(n)$ decays more slowly), so comparisons made at $\alpha T\ll T$ systematically favor simpler models --- the statistical consequence of the folklore ``walk-forward fails P1''.

\begin{remark}[Learning curves and online-to-batch]\label{rem:curves}
Two parts of Assumption~\ref{ass:learning} deserve flags. Monotonicity of $L(n)$ is a genuine restriction: there are learners whose risk increases with the sample size \citep{loog2019}, the double-descent curves of \citet{belkin2019} are non-monotone at the interpolation threshold, \citet{viering2023} review how varied empirical learning curves are, and the universal-learning theory of \citet{bousquet2021} classifies their possible asymptotic shapes. Proposition~\ref{prop:learning} uses only the values of $L$ on $[\alpha T,T]$, so it survives non-monotonicity there with $\min$ and $\max$ of $L$ over that range replacing $L(n_{\max})$ and $L(\alpha T)$. Second, an expanding walk-forward with $\beta T$ blocks of length one is an online learner evaluated prequentially (Remark~\ref{rem:adapted}), and Theorem~\ref{thm:mean} is the combinatorial face of the online-to-batch conversion of \citet{cesabianchi2004}: the average iterate has seen about half of the sample, which is why $\bar\alpha\to1-\beta/2$.
\end{remark}

\subsection{The price of coverage: variance and regime sampling}\label{sec:coverageprice}

\begin{proposition}[Price of coverage]\label{prop:coverage}
(i) \emph{Variance}: if the test-point losses form a stationary sequence with variance $\sigma^2$ and absolutely summable autocorrelations, with long-run variance $\sigma_\infty^2=\sigma^2\bigl(1+2\sum_{d\ge1}\rho(d)\bigr)>0$, then the average loss over the test union $U$ ($|U|=\beta T$) satisfies
\[
\Var\bigl(\hat L\bigr)\ \asymp\ \frac{\sigma_\infty^2}{\beta T}.
\]
(ii) \emph{Regimes (non-stationary worst case)}: take as target the time-averaged risk $\theta=\frac1T\sum_{s\in[T]}r(s)$, with $r(s)$ the risk at time $s$. If two data-generating processes induce the same law for the loss record $(\ell_t)_{t\in U}$ but their risk profiles differ by $\Delta$ on the uncovered part $[T]\setminus U$ and agree on $U$, then any estimator that is a function of the loss record on $U$ incurs worst-case error at least $\frac{(1-\beta)\Delta}{2}$.
\end{proposition}

Part (i) is the standard long-run-variance computation for the mean of a stationary sequence over a set that is a union of long intervals. Its stationarity hypothesis on the losses is exact for a rolling window of fixed length, whose fold models are identically distributed, and approximate for expanding windows, whose fold models change slowly with $|R_i|$. Part (ii) is a two-point argument (Appendix~\ref{app:coverage}). The hypothesis of (ii) is satisfiable: take two processes that coincide on the covered period and on all training data, and differ only in the label-noise level (hence the risk) on an uncovered period that no fold trains or tests on --- for a last-block hold-out with $H$-gap, any change confined to the gap does this. Part (ii) is the deeper face of coverage: the harm of low coverage is not merely variance but \emph{irreparability in the worst case} --- a regime you never tested may hide a $\Delta$, and no statistical technique conjures information out of unobserved intervals. This is why parameters tuned on a late-bull-market hold-out fail in the bear market, and it is the motivation for the multi-path coverage of combinatorial purged cross-validation \citep[CPCV;][]{lopezdeprado2018}.

\subsection{A statistical impossibility theorem}

\begin{corollary}[Statistical impossibility]\label{cor:impossible}
Suppose Assumption~\ref{ass:learning} holds with $L$ convex and strictly decreasing on $[T/2,T]$ (extended to non-integer arguments by linear interpolation), and the test-point losses satisfy the hypotheses of Proposition~\ref{prop:coverage}(i), so that $\Var(\hat L)\ge c\,\sigma_\infty^2/(\beta T)$ for a constant $c>0$. Then for every \emph{strictly causal} scheme with equal-length disjoint test blocks weighted by $|E_i|$,
\[
\MSE\bigl(\hat L;\,L(T)\bigr)\ \ge\ \Bigl[L\bigl((1-\tfrac{\beta}{2})T\bigr)-L(T)\Bigr]^2\ +\ \frac{c\,\sigma_\infty^2}{\beta T}.
\]
The first term is non-decreasing and the second decreasing in $\beta$; the conservation law ($\bar\alpha\le1-\beta/2$, Theorem~\ref{thm:mean}) chains both to the same $\beta$, so no choice of $\beta$ drives both to zero, and the attainable floor $\min_{\beta}\{\cdot\}$ strictly exceeds $c\,\sigma_\infty^2/T$, the variance that the infeasible point $\alpha=\beta=1$ would enjoy. If a non-causal scheme is used to escape this floor, then by Theorem~\ref{thm:main}(a) it must have $\Lambda\ge\alpha+\beta-1$, and by Theorem~\ref{thm:main}(d) it imports a leakage bias of magnitude up to $2M\Leff$, which adds to the learning-curve bias inside the squared term (Theorem~\ref{thm:main}(d) bounds the size of this bias, not its sign; in practice, as in Section~\ref{sec:illustration}, it is optimistic, so it offsets the pessimistic learning-curve bias rather than compounding it).
\end{corollary}

This is the full statistical statement of the impossible trinity: \emph{the three vertices correspond to the three components of the estimation error (leakage bias / learning-curve bias / variance and regime term); the conservation law forbids simultaneously eliminating the last two within strict causality, and the only way to eliminate the first is to pay anti-causal mass priced by the mixing coefficient.}

\subsection{The hardness of the trinity is the memory of the process}\label{sec:memory}

Theorem~\ref{thm:main}(d) writes the price of violating causality as $\bmix(\delta)$, so the hardness of the trinity is set by the memory of the data-generating process. Three regimes:

\paragraph{(a) The i.i.d.\ limit: the trinity vanishes.} If $(W_t)$ is i.i.d.\ and $H=0$, then $\bmix(d)=0$ for all $d\ge1$: anti-causal training does no statistical harm, and $k$-fold, even shuffled, is fully legitimate. The validity of $k$-fold for purely autoregressive models with uncorrelated errors \citep{bergmeir2018} is the model-specific form of this limit: there the augmented process $W_t=(\text{lagged values},Y_t)$ is dependent, and the validity comes from a small algorithm-specific exchange rate $\beta_\Psi$ (Theorem~\ref{thm:ipm}(iii)) rather than from a small $\bmix$. \emph{The trinity is not inherent in the phrase ``time series''; it is created by dependence and non-stationarity.}

\paragraph{(b) Finite memory: causality can be bought back.} If $\bmix(d)\le Ce^{-d/\tau}$ (exponential mixing with memory scale $\tau$; overlapping labels add a plateau for $d<H$), then pushing all anti-causal training beyond margin $h$ caps the harm at $2MCe^{-h/\tau}$, at the sample cost of a sanitization zone of $H$ before and $H+h$ after each test block --- $O\bigl(m(2H+h)/T\bigr)$ in total. This is the mechanism of \textbf{purged $k$-fold + embargo}:
\[
\underbrace{\alpha+\beta\le1+\Lambda}_{\text{(a): }\Lambda\text{ must be bought}}
\qquad\text{while}\qquad
\underbrace{\text{harm}\le 2M\,\bmix(h)}_{\text{(d): the unit price of }\Lambda\text{ is driven to }\approx0}
\]
It does not ``break'' the trinity --- $\Lambda\approx\frac{k-1}{k}$, not one bit less --- but exploits finite memory to \emph{exchange the volume constraint for a severity constraint}. The exchange window is bounded by two rulers: statistically, $h$ must be several multiples of $\tau$ (and the future-side zone must include $H$); combinatorially, the margin is charged to sufficiency (Remark~\ref{rem:horizon}), so the sanitization overhead $k(2H+h)/T$ must stay $\ll1$. When $\tau$ or $H$ becomes comparable to $T/k$, the window closes, and purged $k$-fold degenerates into choosing between ``sanitize everything away'' and ``keep the leak''.

\paragraph{(c) Infinite memory and non-stationarity: not redeemable.} With long memory no affordable $h$ makes $\bmix(h)$ small. Distribution drift is a different failure: it need not raise $\bmix$ at all (an independent but non-identically distributed process has $\bmix(d)=0$), and Theorem~\ref{thm:main}(d) still holds under drift; what fails is the meaning of its reference, since a model that has trained on the future of a test point has seen the regime it is tested in, which the deployed model never will. This regime error, the regime term of Proposition~\ref{prop:coverage}(ii) and the directionality issue of Remark~\ref{rem:symmetry} are not quantities of the mixing framework at all: \emph{an embargo can redeem only the stationary-dependence part of causality; the part demanded by non-stationarity can be paid for in full only by walk-forward-type schemes}, whose report card answers ``re-tuning on a rolling basis by this procedure, could you have made the money at the time'' --- the rehearsal of deployment, including a rehearsal of \emph{the tuning process itself}.

\section{The standard schemes, tuning, and a numerical illustration}\label{sec:schemes}

\subsection{Coordinates}

Table~\ref{tab:coordinates} places the standard schemes in the coordinates $(\alpha,\bar\alpha,\beta,\Lambda,\delta)$, with $k$ folds, walk-forward initial-window (or rolling-window) fraction $w$ and $m$ blocks, purge width $H$, embargo width $h$ (integer rounding ignored; $H_{\mathrm{dep}}$ denotes the dependence scale).

\begin{table}[ht]
\centering
\caption{The standard schemes in the coordinates of the trinity.}
\label{tab:coordinates}
\scriptsize
\setlength{\tabcolsep}{4pt}
\begin{tabular}{lccccc>{\raggedright\arraybackslash}p{3.4cm}}
\toprule
Scheme & $\alpha$ (worst) & $\bar\alpha$ (mean) & $\beta$ & $\Lambda$ & $\delta$ & Effective leakage $\Leff$ \\
\midrule
Last-block hold-out ($\beta_0$) & $1-\beta_0-\frac HT$ & $1-\beta_0-\frac HT$ & $\beta_0$ & $0$ & $+\infty$ & $0$ \\
Expanding walk-forward & $w-\frac HT$ & $1-\frac{m+1}{2m}(1-w)-\frac HT$ & $1-w$ & $0$ & $+\infty$ & $0$ \\
Rolling walk-forward & $w-\frac HT$ & $w-\frac HT$ & $1-w$ & $0$ & $+\infty$ & $0$ \\
$k$-fold (contiguous) & $\frac{k-1}{k}$ & $\frac{k-1}{k}$ & $1$ & $\frac{k-1}{k}$ & $1$ & low--mid: only block-boundary tests have small $\delta_i(t)$; mass $O(kH_{\mathrm{dep}}/T)$ \\
$k$-fold (shuffled) & $\frac{k-1}{k}$ & $\frac{k-1}{k}$ & $1$ & $\approx\frac{k-1}{k}$ & $1$ & \textbf{high}: nearly all tests have $\delta_i(t)=O(1)$ \\
Purged $k$-fold + embargo & \multicolumn{2}{c}{$\frac{k-1}{k}-O(\frac{2H+h}{T})$} & $1$ & $\approx\frac{k-1}{k}$ & $\ge H+h$ & $\le\bmix(h)\approx0$ (needs $h\gtrsim$ memory) \\
\bottomrule
\end{tabular}
\end{table}

Two comparisons in the table are instructive. \emph{Expanding vs.\ rolling walk-forward}: identical in $(\alpha,\beta)$, they differ in $\bar\alpha$ by up to $(1-w)/2$; the rolling window pays this in learning-curve bias (Proposition~\ref{prop:learning}) in exchange for a robustness to drift that the stationary framework cannot price. \emph{Contiguous vs.\ shuffled $k$-fold}: their $(\alpha,\bar\alpha,\beta,\Lambda)$ are identical up to $O(1/T)$, but their pointwise margin distributions differ completely --- the former violates ``at close range'' only at $O(k)$ block boundaries, the latter everywhere. This is the separation of $\Lambda$ and $\Leff$, and Section~\ref{sec:illustration} makes it visible on the same data with the same model.

CPCV reads in these coordinates as follows: by generating $\binom Nk$ block combinations it produces many ``backtest paths'', upgrading the single number $\hat L$ to a distribution of $\hat L$, while purging/embargo keep $\Leff$ controlled --- its main improvement is on the deeper meaning of coverage (sampling of regimes and paths), at the cost of further reduced sufficiency and more computation.

\subsection{The multiplier effect of tuning}\label{sec:tuning}

Everything above evaluates a \emph{fixed} algorithm. Tuning takes a maximum (or minimum) over $N$ configurations, which simultaneously amplifies every edge of the trinity:
\begin{itemize}
\item \textbf{Variance $\to$ selection bias}: even if each configuration's $\hat L$ is unbiased with variance $\varsigma^2$, the selected best carries an optimistic inflation up to about $\varsigma\sqrt{2\ln N}$. Insufficient coverage ($\varsigma\propto1/\sqrt{\beta T}$ by Proposition~\ref{prop:coverage}) enters directly: \emph{low coverage $\times$ many configurations $=$ backtest overfitting}, the phenomenon quantified by the probability of backtest overfitting (PBO) and the deflated Sharpe ratio \citep{bailey2014deflated,bailey2017pbo,white2000}.
\item \textbf{Leakage amplified by selection}: the bias of Theorem~\ref{thm:main}(d) is per configuration; taking a max preferentially selects the configuration \emph{best at exploiting the leak}, pushing the realized optimism toward the upper envelope of per-configuration leakage biases.
\item \textbf{Sufficiency $\to$ ranking distortion}: as in Section~\ref{sec:sufficiency}, learning-curve slopes differ across configurations, so the ranking at small $n$ need not be the ranking at $n=T$.
\end{itemize}
Tuning must therefore be nested: an inner layer (possibly purged $k$-fold) selects parameters, and an outer, strictly causal segment never touched by any selection confirms; reported metrics should be corrected for multiplicity in $N$. The outer segment is the hold-out of adaptive data analysis: \citet{dwork2015} show that it can be queried repeatedly if answers are released through a differentially private mechanism, the selection bias being governed by the information the analyst extracts rather than by $N$ alone.

\subsection{A numerical illustration of volume versus severity}\label{sec:illustration}

The theorems need no experimental confirmation, but the distinction between the anti-causal mass $\Lambda$ and the effective leakage $\Leff$ is the paper's least intuitive point and can be made vivid in a setting where the truth is known exactly.

\textbf{Design.} $T=3000$. Returns $r_t\sim\mathcal N(0,1)$ i.i.d.; labels $Y_t=\sum_{j=1}^Hr_{t+j}$ with $H=20$ (forward $H$-period return): \emph{pure noise, true predictability exactly zero}. Features: four EMAs of returns with half-lives $5/10/20/40$ --- past information only, but serially smooth, so temporal neighbors are feature-space neighbors. Model: 1-NN regression (predict the label of the nearest training point in feature space). Metric: correlation between prediction and realized label (information coefficient, IC). By Remark~\ref{rem:overlap} the label-dependence envelope is $\rho(d)=\frac{(H-d)_+}{H}$. Results are averaged over 8 seeds.

\textbf{Severity sweep.} Training on $\{s:|s-t|>g\}$ for each test point $t$ makes the two-sided margin $\dist(t,R_i)$ of Definition~\ref{def:causal} exactly $g+1$ at every test point --- a direct sweep of the severity variable of P3 (this is hv-block cross-validation with a varying block radius):

\begin{center}
\small
\begin{tabular}{lccccccccc}
\toprule
$g$ & 0 & 2 & 4 & 6 & 8 & 10 & 12 & 14 & $\ge16$ \\
\midrule
reported IC & $+0.340$ & $+0.134$ & $+0.071$ & $+0.040$ & $+0.023$ & $+0.012$ & $+0.004$ & $-0.000$ & $\approx0$ \\
\bottomrule
\end{tabular}
\end{center}

\begin{figure}[tb]
\centering
\includegraphics[width=0.72\textwidth]{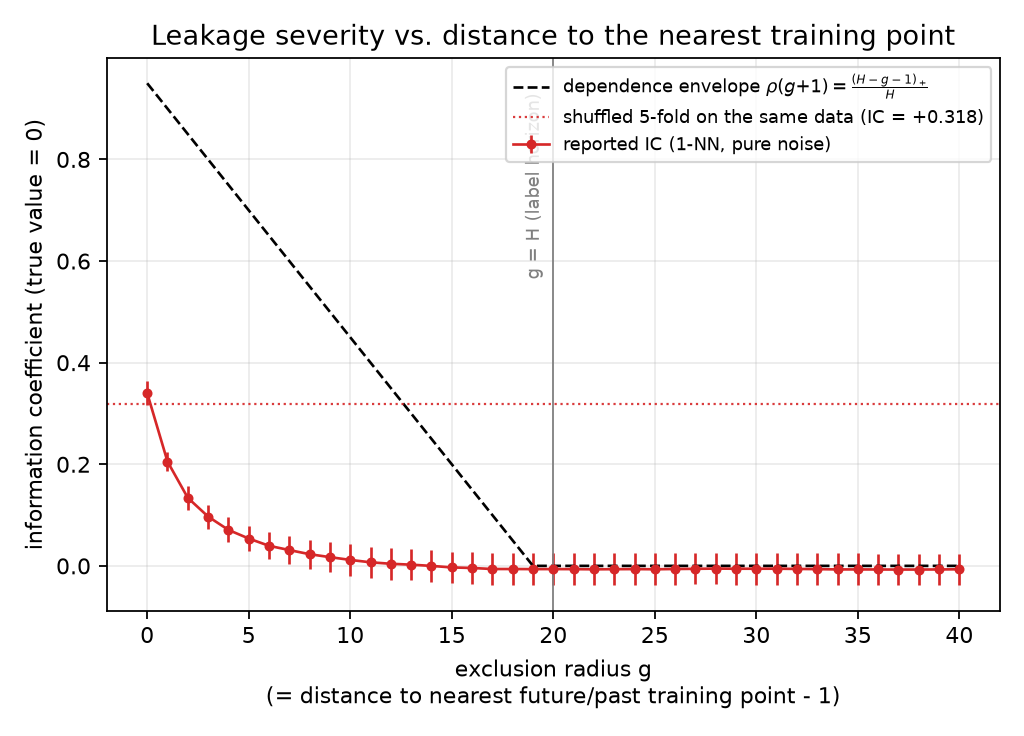}
\caption{Spurious information coefficient of a 1-NN model on pure noise as a function of the exclusion radius $g$ (distance to the nearest training point minus one), with the label-dependence envelope of Remark~\ref{rem:overlap}.}
\label{fig:leakage}
\end{figure}

\noindent Spurious skill decays monotonically with distance and reaches statistical zero before $g\approx H$ (the empirical curve sits below the dependence envelope because the nearest neighbor does not always sit at distance exactly $g+1$; Figure~\ref{fig:leakage}).

\textbf{Named schemes on the same data} (true value $0$):

\begin{center}
\small
\begin{tabular}{lc}
\toprule
Scheme & Reported IC ($\pm$ s.e.) \\
\midrule
Shuffled 5-fold & $\mathbf{+0.318\pm0.012}$ \\
Contiguous 5-fold & $+0.004\pm0.012$ \\
Purged 5-fold (gap $=H$) & $+0.000\pm0.012$ \\
Walk-forward (gap $=H$) & $+0.017\pm0.015$ \\
Last-20\% hold-out (gap $=H$) & $+0.044\pm0.028$ \\
\bottomrule
\end{tabular}
\end{center}

\noindent The walk-forward is expanding, with the last half of the sample as ten equal test blocks (initial window $w=0.5$) and an $H$-gap before each block; the hold-out tests on the last 20\% with the same gap; purged 5-fold deletes $H$ samples on each side of the test block. Shuffled and contiguous 5-fold have the same $(\alpha,\beta,\Lambda)=(0.8,1,0.8)$ (up to $O(1/T)$ for the shuffled $\Lambda$) and differ only in where the anti-causal mass sits: adjacent to every test point in the first case, adjacent to $O(k)$ block boundaries in the second. The reported spurious skill differs by two orders of magnitude --- \emph{volume is harmless, proximity is harmful}. The hold-out, with the smallest $\beta$, shows the largest standard error, the variance price of Proposition~\ref{prop:coverage}(i).

\section{Practical guidance}\label{sec:practice}

The paper compressed into an operating card:
\begin{enumerate}
\item \textbf{Measure memory first, choose the scheme second.} Estimate two rulers: the label horizon $H$ (known from the label definition) and the dependence scale $\tau$ of features/losses (autocorrelation half-life; for financial data, remember volatility clustering). All gaps/purges/embargoes should be denominated in $H+c\tau$ with $c\approx2$--$3$.
\item \textbf{Stationarity credible, goal is model comparison/tuning} $\to$ purged $k$-fold + embargo (or CPCV): it buys $\alpha,\beta\approx1$ at minimal volume $\Lambda$ and effective leakage $\le\bmix(h)$ (Section~\ref{sec:memory}(b)). Check that the sanitization overhead $k(2H+h)/T\ll1$; otherwise reduce $k$.
\item \textbf{Non-stationarity feared, or ``expected live performance'' to be quoted} $\to$ expanding walk-forward (it \emph{is} the causal frontier, Corollary~\ref{cor:frontier}); report the \emph{per-fold-versus-training-size} curve rather than a single average (the average carries the learning-curve bias of Proposition~\ref{prop:learning}; the per-fold curve extrapolates toward $L(T)$); rehearse the tuning process inside the rolling procedure.
\item \textbf{Tuning must be nested}: inner selection may use item~2; an outer strictly causal segment never touched by selection confirms. Deflate the winner of $N$ configurations at the $\varsigma\sqrt{2\ln N}$ scale, or use the deflated Sharpe ratio / PBO directly.
\item \textbf{Report the coordinates} $(\alpha,\bar\alpha,\beta,\Lambda,\delta)$ of your scheme: one line states where your backtest stands on the trinity and what it paid. A referee or risk manager needs only the ratio of $\delta$ to $H+\tau$ to decide whether to trust your $\Lambda>0$.
\item \textbf{Red line}: shuffled $k$-fold is unusable on any data with serial dependence or overlapping labels (Section~\ref{sec:illustration}); for a ``too-good backtest'', the first suspect is always a small-$\delta$ leak, never alpha.
\end{enumerate}

\section{Related work}\label{sec:related}

\paragraph{Validation schemes for dependent data.} The h-block cross-validation of \citet{burman1994} and the hv-block cross-validation of \citet{racine2000} first formalized ``delete a neighborhood of the test point'' as the dependent-data correction of cross-validation; the exclusion-radius sweep of Section~\ref{sec:illustration} is hv-block with a varying radius. \citet{bergmeir2012} and \citet{bergmeir2018} established the validity of $k$-fold cross-validation for purely autoregressive models with uncorrelated errors, the model-specific instance of Section~\ref{sec:memory}(a). \citet{tashman2000} codifies the rolling-origin (walk-forward) tradition; \citet{cerqueira2020} and \citet{arian2024} compare the standard schemes empirically, the latter including CPCV in a synthetic controlled environment. \citet{lopezdeprado2018} systematized overlapping-label leakage, purged $k$-fold, embargo, and CPCV; Theorem~\ref{thm:main}(d) is a theorem-ization of the embargo prescription (the unit of embargo width is the decay scale of the mixing coefficients, and the pointwise margin $\delta_i(t)$ is a sufficient statistic for the bound on the harm), while Theorem~\ref{thm:main}(b) states what the embargo costs in sufficiency. Our contribution is not another scheme but the proof that all schemes obey one inequality, together with its Pareto frontier and its violation price list.

\paragraph{Learning theory for dependent processes.} The proof of Theorem~\ref{thm:main}(d) rests on the coupling lemma of \citet{berbee1979} and the blocking inequality of \citet{yu1994}. The closest theoretical neighbour is the work of \citet{kuznetsov2015,kuznetsov2020}, who prove generalization bounds for forecasting non-stationary mixing processes via mixing coefficients and a discrepancy measure. Their object is the generalization of a trained predictor, ours is the design space of validation schemes, and the two are complementary: their discrepancy quantifies the non-stationarity that Section~\ref{sec:memory}(c) declares non-redeemable, and Theorem~\ref{thm:ipm} shows that the same discrepancy is the model-specific exchange rate of the causality vertex. On the multiplier effect of tuning (Section~\ref{sec:tuning}), the Reality Check of \citet{white2000} and the deflated Sharpe ratio and probability of backtest overfitting of \citet{bailey2014deflated,bailey2017pbo} quantify the selection inflation that low coverage feeds.

\section{Conclusion}\label{sec:conclusion}

We proved that the three demands of time-series validation --- training sufficiency, test coverage, temporal causality --- obey a single inequality, $\alpha+\min\{\beta,\delta/T\}\le1$, that is independent of all statistical assumptions; that its causal face $\alpha+\beta\le1$ has the walk-forward family as its exact Pareto frontier; that the only exit is anti-causal training mass, at a combinatorial price of $\Lambda\ge\alpha+\beta-1$, placed within $(1-\alpha)T$ of the tests so that distance is paid out of sufficiency; and that the statistical harm of that mass is denominated in the distance from the test point to the first future training point, converted at the mixing coefficients of the process. The deep principle is not complicated: \emph{time is totally ordered and information flows along it in one direction; any scheme that tries to both ``use up the past'' and ``test all of the past'' must place some training data in the near future of some test data; and the harm of future data is a function of its distance to the test point, at an exchange rate set by the memory of the process.} The trinity cannot be had whole --- but the price of each edge can be written down, measured, and minimized. That is what this paper hopes to supply: not one more splitting scheme, but the common ledger of all of them.

Only the combinatorial parts of the paper use the fact that time is a line. When dependence lives on a graph --- spatial, panel, network or phylogenetic data, for which \citet{roberts2017} survey blocked cross-validation --- the margin becomes graph distance from a test node to the nearest training node, the mixing coefficient a decay of correlations on the graph, Theorem~\ref{thm:ipm} applies verbatim, and Theorem~\ref{thm:main}(b) becomes a vertex-isoperimetric statement: keeping the tests at distance $\delta$ from the training data sterilizes the $\delta$-neighbourhood of the test set, of size $2\delta$ on the line, of order $\delta^d$ on a $d$-dimensional grid, and exponential in $\delta$ on an expander. The growth rate of the dependence graph is thus the exchange rate between distance and sufficiency, and the line is the cheapest case.

\appendix

\section{Proofs}\label{app:proofs}

Proofs that are immediate from the statements (Corollaries~\ref{cor:conservation} and~\ref{cor:frontier}, Proposition~\ref{prop:learning}, Proposition~\ref{prop:coverage}(i)) are indicated in the text and omitted here.

\subsection{Theorem \ref{thm:main}(a)--(c)}\label{app:main}

\emph{(a)} Let $U=\bigcup_iE_i$ and $p=\min U$. From $U\subseteq[p,T]$,
\begin{equation}\label{eq:pbound}
\beta T\ \le\ T-p+1,\qquad\text{with equality iff }U=[p,T].
\end{equation}
Pick a fold $i$ with $p\in E_i$; then $p=\min E_i$, and by Definition~\ref{def:causal} $\Lambda_iT=|R_i\cap(p,T]|$ exactly (note $p\notin R_i$). Hence
\begin{equation}\label{eq:chain}
\Lambda T\ \ge\ \Lambda_iT\ =\ |R_i|-\bigl|R_i\cap[1,p-1]\bigr|\ \ge\ \alpha T-(p-1)\ \ge\ \alpha T-(1-\beta)T,
\end{equation}
where the three inequalities are equalities iff, respectively, $\Lambda_i=\Lambda$; $|R_i|=\alpha T$ and $R_i\supseteq[1,p-1]$; and \eqref{eq:pbound} is an equality. This proves $\alpha+\beta\le1+\Lambda$ together with the stated equality characterization.

\emph{(b)} Let fold $i$ have $\Lambda_i>0$; then $t=\min E_i$ has a training point after it, so $\delta_i(t)<\infty$. The $\delta_i(t)$ indices $t,\,t+1,\dots,t+\delta_i(t)-1$ all lie in $[T]$ and none is in $R_i$ ($t\notin R_i$ by disjointness, the others by minimality of $\delta_i(t)$), so they are $\delta_i(t)$ distinct elements of $[T]\setminus R_i$, a set of size $T-|R_i|$. Hence $\delta_i\le\delta_i(t)\le T-|R_i|$. If $\Lambda>0$, some fold qualifies and $|R_i|\ge\alpha T$ gives $\delta\le\delta_i\le(1-\alpha)T$; finally, $\alpha+\beta>1$ forces $\Lambda\ge\alpha+\beta-1>0$ by (a).

\emph{(c)} If $\scheme$ is strictly causal, $\Lambda=0$ and (a) gives $\alpha+\beta\le1$, which is the claim since $\min\{\beta,+\infty\}=\beta$. Otherwise $\Lambda>0$, and (b) gives $\alpha+\delta/T\le1$, so $\alpha+\min\{\beta,\delta/T\}\le\alpha+\delta/T\le1$. Equality on the causal frontier is attained by the expanding walk-forward construction of Corollary~\ref{cor:frontier}, whose worst fold is $R_1=[1,(1-\beta)T]$. \qed

\subsection{Theorem \ref{thm:main}(d)}\label{app:severity}

Fix fold $i$ and test point $t$; write $A=W_{R^-}$, $B=W_t$, $C=W_{R^+}$, so that $\ell_t=\psi(A,B,C)$ for some measurable $\psi$ with $|\psi|\le M$ (any internal randomness of the algorithm can be absorbed into $\psi$ by conditioning). By the definition of the margins, $(A,B)$ is $\sigma(W_s:s\le t)$-measurable, $C$ is $\sigma(W_s:s\ge t+\delta^+)$-measurable, $A$ is $\sigma(W_s:s\le t-\delta^-)$-measurable and $B$ is $\sigma(W_s:s\ge t)$-measurable. We use two standard facts. First, for probability measures $P,Q$ and $|\psi|\le M$, $|\E_P\psi-\E_Q\psi|\le 2M\tv{P-Q}$. Second, Berbee's coupling lemma \citep[see][Ch.~1]{berbee1979,doukhan1994}: for random elements $X,Y$ with $\tv{\law(X,Y)-\law(X)\otimes\law(Y)}\le\epsilon$, there exists on an enlarged probability space a copy $Y^*$ of $Y$ independent of $X$ with $\Prob(Y^*\neq Y)\le\epsilon$.

\emph{Future side.} Assumption~\ref{ass:mixing} with $a=t$, $d=\delta^+$ gives $\tv{\law((A,B),C)-\law(A,B)\otimes\law(C)}\le\bmix(\delta^+)$ (trivially so when $\delta^+=+\infty$, i.e.\ $R^+=\varnothing$). Berbee's lemma yields $C^*$, equal in law to $C$, independent of $\sigma(W_s:s\le t)$, with $\Prob(C^*\neq C)\le\bmix(\delta^+)$; and $\tilde\ell_t=\psi(A,B,C^*)$ satisfies
\[
\bigl|\E[\ell_t]-\E[\tilde\ell_t]\bigr|
\ \le\ \E\bigl|\psi(A,B,C)-\psi(A,B,C^*)\bigr|
\ \le\ 2M\,\Prob(C^*\neq C)
\ \le\ 2M\,\bmix(\delta^+).
\]

\emph{Both sides.} Let $P=\law(A,B,C)$ and $Q=\law(A)\otimes\law(B)\otimes\law(C)$. By the triangle inequality,
\begin{align*}
\tv{P-Q}\ \le\ &\tv{\law(A,B,C)-\law(A,B)\otimes\law(C)}\\
&+\ \tv{\law(A,B)\otimes\law(C)-\law(A)\otimes\law(B)\otimes\law(C)}.
\end{align*}
The first term is at most $\bmix(\delta^+)$ as above. The second equals $\tv{\law(A,B)-\law(A)\otimes\law(B)}$ (tensoring with a common factor does not change total variation), which is at most $\bmix(\delta^-)$ by Assumption~\ref{ass:mixing} with $a=t-\delta^-$, $d=\delta^-$. Since $\E[\ell_t]=\E_P\psi$ and $\E[\ell^\circ_t]=\E_Q\psi$, the first standard fact gives $|\E[\ell_t]-\E[\ell^\circ_t]|\le2M[\bmix(\delta^-)+\bmix(\delta^+)]$. (Cf.\ the blocking inequality of \citet[Lemma~4.1]{yu1994} for three blocks; constants differ with the normalization of total variation.)

Averaging over $t\in E_i$ gives the fold-level inequalities. \qed

\subsection{Theorem \ref{thm:mean} (mean sufficiency)}\label{app:mean}

Let the test sets be disjoint intervals of equal length $b=\beta T/m$, sorted by left endpoint as $E_{(1)},\dots,E_{(m)}$. For the $j$-th: the intervals $E_{(j)},E_{(j+1)},\dots,E_{(m)}$ are disjoint and contained in $[\min E_{(j)},\,T]$, hence $(m-j+1)\,b\le T-\min E_{(j)}+1$, i.e.\ $\min E_{(j)}-1\le T-(m-j+1)b$. Strict causality gives $|R_{(j)}|\le\min E_{(j)}-1$. Summing,
\[
\sum_{j=1}^m|R_{(j)}|\ \le\ mT-b\sum_{j=1}^m(m-j+1)\ =\ mT-b\,\frac{m(m+1)}{2},
\]
and dividing by $mT$ yields $\bar\alpha=\frac1{mT}\sum_j|R_{(j)}|\le1-\frac{m+1}{2m}\beta$ (recall $w_i=1/m$). In the construction of Corollary~\ref{cor:frontier} with equal blocks, $|R_{(j)}|=\min E_{(j)}-1$ and the blocks abut and end at $T$, so every inequality is tight. \qed

\subsection{Proposition \ref{prop:coverage}(ii) (regime term)}\label{app:coverage}

Let $P_0,P_1$ be the two processes, with risk profiles $r_0,r_1$ satisfying $r_1(s)-r_0(s)=\Delta$ for $s\notin U$ and $0$ for $s\in U$, so that the targets satisfy $\theta_1-\theta_0=(1-\beta)\Delta$. Any estimator $\hat\theta=\hat\theta\bigl((\ell_t)_{t\in U}\bigr)$ has the same distribution under $P_0$ and $P_1$, hence
\[
\max_{j\in\{0,1\}}\E_{P_j}\bigl|\hat\theta-\theta_j\bigr|
\ \ge\ \tfrac12\Bigl(\E_{P_0}\bigl|\hat\theta-\theta_0\bigr|+\E_{P_0}\bigl|\hat\theta-\theta_1\bigr|\Bigr)
\ \ge\ \frac{|\theta_1-\theta_0|}{2}=\frac{(1-\beta)\Delta}{2}. \qquad\qed
\]

\subsection{Corollary \ref{cor:impossible} (statistical impossibility)}\label{app:impossible}

By Definition~\ref{def:alpha}, $\bar n=\sum_iw_i|R_i|=\bar\alpha T$, and with equal-length disjoint test blocks weighted by $|E_i|$ Theorem~\ref{thm:mean} gives $\bar n\le(1-\frac{m+1}{2m}\beta)T\le(1-\frac\beta2)T$. With $L$ convex, Proposition~\ref{prop:learning} gives bias $\ge L(\bar n)-L(T)\ge L\bigl((1-\frac\beta2)T\bigr)-L(T)\ge0$; Proposition~\ref{prop:coverage}(i) gives $\Var(\hat L)\ge c\sigma_\infty^2/(\beta T)$. Since $\MSE=\text{bias}^2+\Var$, the displayed bound follows. Monotonicity in $\beta$ of the two terms is clear. For the floor: at $\beta=1$ the first term is $[L(T/2)-L(T)]^2>0$ because $L$ is strictly decreasing on $[T/2,T]$, so the sum exceeds $c\sigma_\infty^2/T$; for $\beta<1$ the second term alone exceeds $c\sigma_\infty^2/T$. Since $\beta T$ is an integer, $\beta$ ranges over a finite set and the minimum is attained. The third term for non-causal schemes follows from Theorem~\ref{thm:main}(a) combined with Theorem~\ref{thm:main}(d). \qed

\end{document}